\documentclass[sigconf]{acmart}

\AtBeginDocument{%
  \providecommand\BibTeX{{%
    \normalfont B\kern-0.5em{\scshape i\kern-0.25em b}\kern-0.8em\TeX}}}

\copyrightyear{2020}
\acmYear{2020}
\setcopyright{acmcopyright}
\acmConference[MM '20]{Proceedings of the 28th ACM
	International Conference on Multimedia}{October 12--16, 2020}{Seattle, WA, USA}
\acmBooktitle{Proceedings of the 28th ACM International Conference on Multimedia
	(MM '20), October 12--16, 2020, Seattle, WA, USA}
\acmPrice{15.00}
\acmDOI{10.1145/3394171.3413970}
\acmISBN{978-1-4503-7988-5/20/10}

\usepackage{epsfig}
\usepackage{graphicx}
\usepackage{color, colortbl}
\usepackage{multirow}
\usepackage{enumitem}
\usepackage{amsmath}
\usepackage{makecell}
\usepackage{hhline}
\usepackage{cellspace}
\definecolor{mygray}{gray}{.92}
\definecolor{LightCyan}{rgb}{0.88,1,1}
\usepackage{bbding}
\usepackage{pifont}
\usepackage{wasysym}
\usepackage[utf8]{inputenc}
\usepackage{cleveref}
\crefname{section}{§}{§§}
\Crefname{section}{§}{§§}
\newcommand{\x}[1]{{\mathbf #1}}
\newcommand{\myparagraph}[1]{{\vspace{0.5em} \noindent \bf #1}}

\definecolor{mypink}{rgb}{1.0, 0.0, 0.56}
\begin{document}

\title{Forest R-CNN: Large-Vocabulary Long-Tailed Object Detection and Instance Segmentation}

\author{Jialian Wu$^{1}$, Liangchen Song$^{1}$, Tiancai Wang$^{2}$, Qian Zhang$^{3}$ and Junsong Yuan$^{1}$}
\affiliation{\institution{$^1$State University of New York at Buffalo\\ $^2$Tianjin University \qquad \quad $^3$Horizon Robotics, Inc.}}
\email{{jialianw,jsyuan}@buffalo.edu}

\renewcommand{\shortauthors}{Jialian Wu, et al.}

\begin{abstract}
Despite the previous success of object analysis, detecting and segmenting a large number of object categories with a long-tailed data distribution remains a challenging problem and is less investigated. For a large-vocabulary classifier, the chance of obtaining noisy logits is much higher, which can easily lead to a wrong recognition. In this paper, we exploit prior knowledge of the relations among object categories to cluster fine-grained classes into coarser parent classes, and construct a classification tree that is responsible for parsing an object instance into a fine-grained category via its parent class. In the classification tree, as the number of parent class nodes are significantly less, their logits are less noisy and can be utilized to suppress the wrong/noisy logits existed in the fine-grained class nodes. As the way to construct the parent class is not unique, we further build multiple trees to form a classification forest where each tree contributes its vote to the fine-grained classification. To alleviate the imbalanced learning caused by the long-tail phenomena, we propose a simple yet effective resampling method, NMS Resampling, to re-balance the data distribution. Our method, termed as Forest R-CNN, can serve as a plug-and-play module being applied to most object recognition models for recognizing more than $1000$ categories. Extensive experiments are performed on the large vocabulary dataset LVIS. Compared with the Mask R-CNN baseline, the Forest R-CNN significantly boosts the performance with $11.5\%$ and $3.9\%$ AP improvements on the rare categories and overall categories, respectively. Moreover, we achieve state-of-the-art results on the LVIS dataset. \emph{Code is available at \textcolor{mypink}{\url{https://github.com/JialianW/Forest_RCNN}.}}
\vspace{-1mm}
\end{abstract}



\maketitle
{\fontsize{8pt}{8pt}\selectfont
	\textbf{ACM Reference Format:}\\
	Jialian Wu, Liangchen Song, Tiancai Wang, Qian Zhang and Junsong Yuan. 2020. Forest R-CNN: Large-Vocabulary Long-Tailed Object Detection and Instance Segmentation. In \textit{Proceedings of the 28th ACM International Conference on Multimedia (MM '20), October 12--16, 2020, Seattle, WA, USA.} ACM, New York, NY, USA, 9 pages. https://doi.org/10.1145/3394171.3413970}

\section{Introduction}

\begin{figure}
	\centering
	\includegraphics[width=1\linewidth]{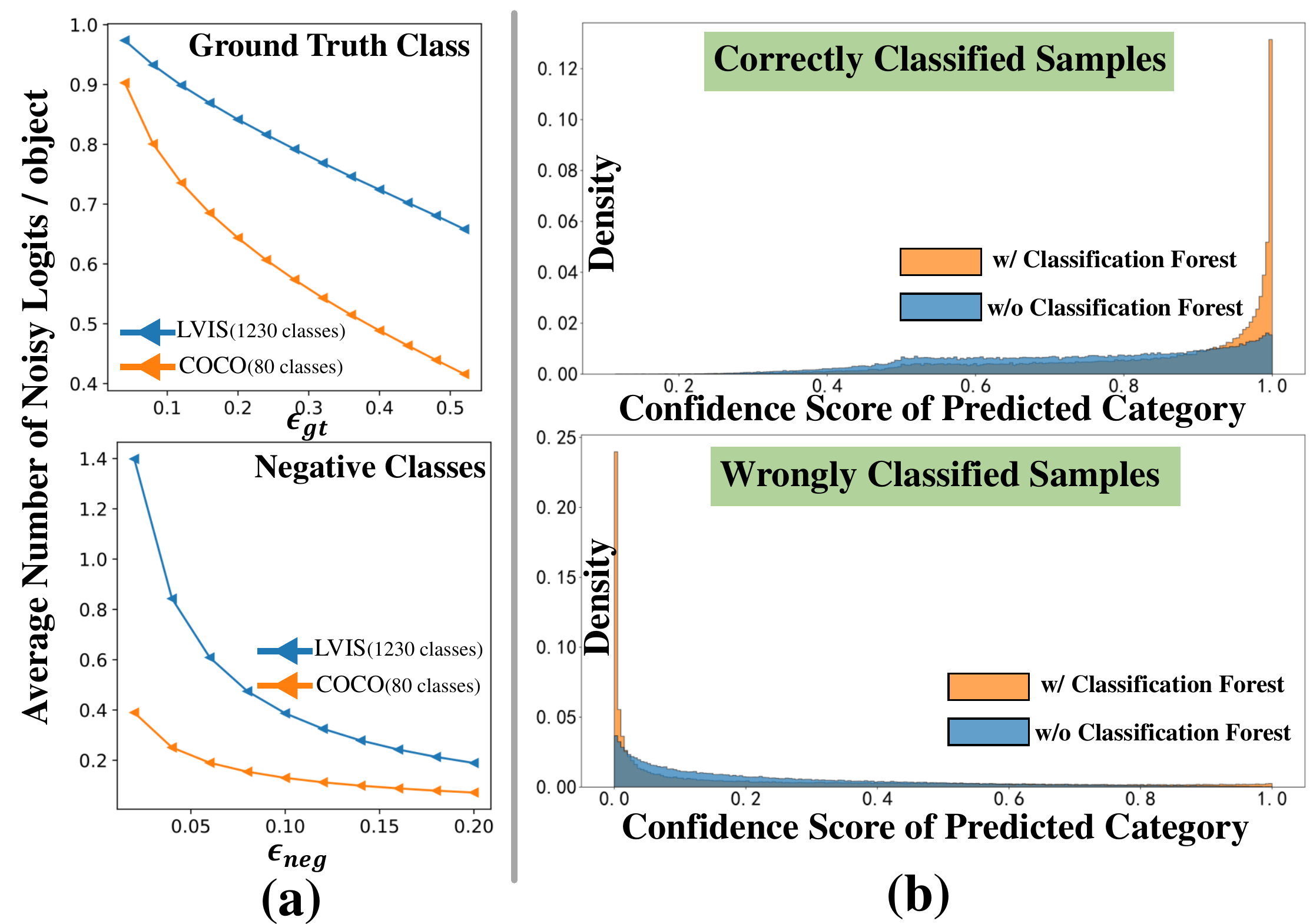}
	\vspace{-7mm}
	\caption{(a) Statistics of noisy logits of COCO \texttt{val} set \emph{vs.} LVIS~\cite{lvis} \texttt{val} set tested by Mask R-CNN with ResNet-50-FPN where the noisy logit is defined in Eq.~\ref{eqn:noisy}. It is seen that the large vocabulary brings a higher chance of noisy logits which can easily lead to a wrong recognition. In the worst case, an object can have $1$ and $N-1$ noisy logits for ground truth class and negative classes, respectively, where $N$ is the total number of classes. (b) Probability density distribution of confidence scores on the LVIS v0.5 \texttt{val} set. With our proposed classification forest, the confidence scores of correctly classified samples are improved (top) and those of wrongly classified samples are suppressed (bottom). The top figure is measured by considering the background class.}
	\label{fig:fig1}
	\vspace{-3mm}
\end{figure}

With the renaissance of deep convolutional neural networks (CNNs), recent years have witnessed great progress in object recognition, including object detection~\cite{rcnn,faster_rcnn,snip,ssd,li_cvpr20,pang_tcb16} and instance segmentation~\cite{maskrcnn,mask_score,yolact,dai_eccv16,xu_iccv19}. Object recognition plays a central role in visual learning and has been widely applied to various applications, \emph{e.g.,} person retrieval~\cite{xiao_cvpr18}, human-object interaction~\cite{tiancai_cvpr20}, and visual reasoning~\cite{jiajun_nips17}.

Most great works on object recognition have been thoroughly investigated in the few category regime, \emph{e.g.,} PASCAL VOC~\cite{pascal_voc} (20 classes) and COCO~\cite{coco} (80 classes). However, practical applications are urging the need for recognizing a large number of categories with a long-tailed data distribution, which is a great challenge for most existing methods. Generally, the challenge boils down to two aspects: \textbf{(i)} As the number of categories grows, the chance of obtaining noisy logits in a classifier becomes higher (\emph{i.e.,} inaccurate classifier predictions on either the ground truth class or other negative classes) as shown in Fig.~\ref{fig:fig1} (a). This increases the difficulty for generating a correct category label or a high confidence score on the ground truth category, therefore easily leading to a wrong recognition. \textbf{(ii)} The long-tail phenomena inherently occur in a large vocabulary scenario and cause extreme imbalanced data distribution, where few classes (\emph{a.k.a.} head class) appear very often yet most of other classes (\emph{a.k.a.} tail class) rarely appear. Due to the imbalanced distribution, most tail classes are overwhelmed by head classes in training, making it difficult to well learn effective classifier and feature representations especially for tail classes.

\begin{figure}
	\centering
	\includegraphics[width=1\linewidth]{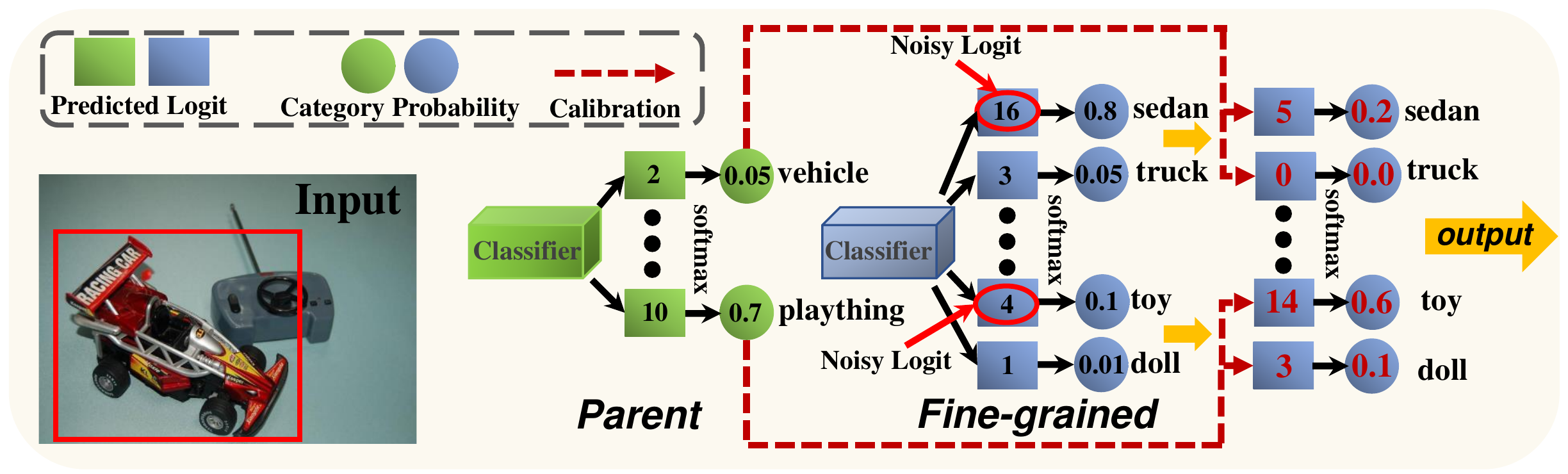}
	\vspace{-7mm}
	\caption{Brief illustration of the classification tree.}
	\label{fig:illustration}
	\vspace{-4mm}
\end{figure}

In this paper, we propose a novel classification forest together with a simple yet effective data resampling method, striving to alleviate the above problems \textbf{(i)} and \textbf{(ii)}, respectively. For \textbf{(i)}, we exploit prior knowledge of the relations among categories to cluster thousands of fine-grained classes into tens of parent classes, and construct a classification tree that is responsible for parsing an object into a fine-grained node via its parent node. Since the number of parent classes is significantly less, the parent classifier within the tree obtains fewer noisy logits. We then utilize the parent class probabilities estimated from the parent classifier to suppress the noisy logits produced by the fine-grained classifier. Moreover, in terms of different types of prior knowledge, we construct multiple classification trees to form a classification forest, where each tree will contribute its vote to the fine-grained classification. To illustrate the idea of the classification tree, we show an example in Fig.~\ref{fig:illustration} where the fine-grained classifier wrongly predicts a ``toy'' as a ``sedan'' with $p(sedan)=0.8$ while the parent class probability of ``sedan'' is $p(vehicle)=0.05$. In the classification tree, the noisy logit of class ``sedan'' will be calibrated and suppressed by $p(vehicle)$. By suppressing the noisy logits, our method is able to improve the confidence scores of ground truth class and suppress those of other negative classes as shown in Fig.~\ref{fig:fig1} (b). For \textbf{(ii)}, we propose a data resampling method, named as NMS Resampling, to adaptively adjust the Non-maximum Suppression (NMS) threshold for different categories based on their category frequency in training. The NMS Resampling can re-balance the data distribution by preserving more training proposal candidates from tail classes while suppressing those from head classes in the bounding box NMS procedure.

Recent attempts~\cite{equalization_loss,tan2020equalization,rfcn3000,yolo9000,ouyang_cvpr16} also strive to recognize objects under the setting of large vocabulary. For example, ~\cite{equalization_loss} studies the influence of loss functions on head classes and tail classes, and propose an equalization loss to reduce the negative influence of loss gradients on tail classes. Note that~\cite{equalization_loss} is complementary to our method and can be employed together for gaining performance. Other works~\cite{rfcn3000,yolo9000,ouyang_cvpr16} also utilize a hierarchical tree structure for better classification, which however may still easily yield a wrong recognition due to inaccurate parent class nodes generated by a single tree. Our method, instead, exploits different types of prior knowledge to build a forest, in which classification is achieved in the form of plurality vote.

\myparagraph{Contributions:} In this work, we propose a novel classification forest that incorporates relations among fine-grained categories via different prior knowledge. When dealing with large vocabulary object recognition, our classification forest can better perform via suppressing the noisy logits existed in the fine-grained classifier. In addition, the NMS Resampling method is proposed for re-balancing the data distribution of a long-tailed dataset during training. Our method, termed as Forest R-CNN, is exhaustively evaluated on the large vocabulary object recognition dataset LVIS~\cite{lvis} which contains more than $1000$ categories. The Forest R-CNN achieves significant AP gains of $11.5\%$ and $3.9\%$ on the rare categories and overall categories, respectively.

\vspace{-2mm}
\section{Related Work}
\vspace{-2mm}
\myparagraph{Object Detection and Instance Segmentation.} Object detection, which plays an important role in various vision applications, has been actively investigated in past decades~\cite{faster_rcnn,focal_loss,SML,TFAN,snip,cao_tcsvt19,cao_cvpr20,cao_cvpr19}. With deep learning, two mainstream frameworks, \emph{i.e.,} single-stage detector~\cite{ssd,focal_loss,cao_iccv19} and two-stage detector~\cite{rcnn,fast_rcnn,faster_rcnn}, have dramatically improved both accuracy and efficiency. Instance segmentation can be viewed as an extension of object detection, where each object instance is bounded by a precise mask instead of a rough bounding box. As a pioneer work, Mask R-CNN~\cite{maskrcnn} achieves instance segmentation by building an extra segmentation head upon the two-stage object detector Faster R-CNN~\cite{faster_rcnn}. On the basis of the two-stage fashion,~\cite{fcis,mask_score,panet} further propose several framework modifications so as to yield better performance. On the other hand, single-stage approaches~\cite{yolact,xu_iccv19,dai_eccv16,cao_eccv20}, aiming at a faster inference speed, adopt a more straightforward way to directly generate instance masks from the whole feature maps. Different from the previous works that mostly perform in the few category regime, our method focuses more on object detection and instance segmentation in a large vocabulary with long-tailed data distribution.

\myparagraph{Large-Vocabulary and Long-Tailed Visual Recognition.} To be applied in the natural world, a visual recognition model is expected to deal with a large number of categories with the long-tail phenomena. One prominent method for long-tailed visual recognition is data re-balancing, which is usually divided into two groups: \emph{re-sampling}~\cite{cls_sample,resample2,resample4,lst,bbn} and \emph{re-weighting}~\cite{reweight1,cls_balance_loss,reweight3}. \emph{Re-sampling} schemes typically oversample the data from minority classes while undersample those from frequent classes in training. \emph{Re-weighting} schemes, instead, assign larger loss weights for the samples from minority classes in training. In addition to data re-balancing, many great works have been made using different ways, \emph{e.g.,} model fine-tuning~\cite{ouyang_cvpr16,reweight3}, metric learning~\cite{reweight1,zhang_iccv17}, meta-learning~\cite{liu_cvpr19}, and knowledge transfer learning~\cite{zhong_cvpr19}. Recently,~\cite{equalization_loss,ouyang_cvpr16,yolo9000,rfcn3000} also strive to solve the problem of object recognition in a large vocabulary.~\cite{ouyang_cvpr16,yolo9000,rfcn3000} exploit a single hierarchical tree structure to aid the fine-grained classification. Different from the above approaches, we propose a novel classification forest aiming at reducing the noisy logits existed in the fine-grained classifier to improve classification capability in a large vocabulary. Besides, in contrast to most \emph{re-sampling} methods that resample data on the image level, our proposed NMS Resampling scheme re-balances data on the instance level.

\section{Forest R-CNN}

\subsection{Problem Formulation}
\vspace{-1mm}
\label{subsec:problem_formulation}
Given a static image $\x{I} \in  \mathbb{R}^{W \times H\times3}$, an object recognition model is required to perform object detection and instance segmentation simultaneously. Thus, each recognized object instance is associated with four predictions: $\x{b} \in \mathbb{R}^{4}$, $\x{m} \in [0,1]^{W \times H}$, $l \in \mathbb{R}^{1}$, $s \in \mathbb{R}^{1}$, which are bounding box, segmentation mask, category label, and confidence score, respectively. In this paper, we implement our method based on the baseline model Mask R-CNN \cite{maskrcnn}. Let us denote by $\x{f}_{\rm img}=\mathcal{N}_{\rm feat}(\x{I}) \in\mathbb{R}^{\frac{W}{r}\times \frac{H}{r}\times C}$ the feature maps extracted from input image $\x{I}$, where $\mathcal{N}_{\rm feat}$, $r$, and $C$ are the backbone network, feature stride, and feature channels, respectively. A region proposal network (RPN) \cite{faster_rcnn} is built upon $\x{f}_{\rm img}$ to generate a set of proposal candidate boxes $\{\x{b}_{p}\in\mathbb{R}^{4}\}$. For each proposal candidate, its corresponding proposal features $\x{f}_{\rm roi} \in \mathbb{R}^{7\times 7\times C}$ are obtained by $\x{f}_{\rm roi}=\phi(\x{f}_{\rm img}, \x{b}_{p})$, where $\phi$ is the RoI Align operation \cite{maskrcnn}. On the basis of $\x{f}_{\rm roi}$, three head networks, $\mathcal{N}_{\rm cls}$, $\mathcal{N}_{\rm box}$, and $\mathcal{N}_{\rm mask}$ are employed to generate $s,l=\mathcal{N}_{\rm cls}(\x{f}_{\rm roi})$, $\x{b}=\mathcal{N}_{\rm box}(\x{f}_{\rm roi})$, and $\x{m}=\mathcal{N}_{\rm mask}(\x{f}_{\rm roi})$, in which $\mathcal{N}_{\rm cls}$ and $\mathcal{N}_{\rm box}$ share part of network parameters. 

$\mathcal{N}_{\rm cls}$ is critical for an object recognition model, since classification result is the premise for evaluating boxes and masks. Let $\{x_{i}\}_{i=1}^{N}$ be the category set of a given dataset, where $N$ is the number of total classes and $x_{i}$ is the $i$-th fine-grained class. $p(x_{i})$ indicates the classifier inferred probability that a given object belongs to class $x_{i}$, and it is calculated by the softmax function $p(x_{i}) = \frac{f_{x_{i}}}{\sum^{N}_{a=1}f_{x_{a}}}$, where $f_{x_{i}}=e^{z_{i}}$ and $z_{i}$ is predicted by the $i$-th neuron in the final layer of $\mathcal{N}_{\rm cls}$. In this paper, we name $f_{x_{i}}$ as the logit of class $x_{i}$ (exponential version) and call $f_{x_i}$ as a noisy logit when:

\vspace{-2mm}
\begin{equation}
\label{eqn:noisy}
\begin{cases}
\frac{f_{x_{i}}}{\sum^{N}_{a=1}f_{x_{a}}} < 1 - \epsilon_{gt}, & \text{if } x_{i} = x_{gt} \\
\frac{f_{x_{i}}}{\sum^{N}_{a=1}f_{x_{a}}} > \epsilon_{neg} ,     & \text{if } x_{i} \neq x_{gt}
\end{cases},
\end{equation}
where $x_{gt}$ is the ground truth class and $0<\epsilon_{gt},\epsilon_{neg} <1$ are two small constants. In~\cite{maskrcnn}, the confidence score of class $x_{i}$ and the category label $l$ of a given object are obtained by:

\vspace{-2mm}
\begin{equation}
\label{eqn:eq_base}
\begin{aligned}
&s_{i}= p(x_{i}),\\
&l=\underset{i}{\operatorname{argmax}}\ s_{i}, \quad i=1,2,...,N.\\
\end{aligned}
\end{equation}

Generally speaking, to yield a correct label and a high confidence score on $x_{gt}$, Eq.~\ref{eqn:eq_base} is satisfactory in the few category regime. However, in a large vocabulary scenario, \emph{e.g.,} $N>1000$, there exist many more noisy logits within $\{f_{x_{i}}\}_{i=1}^{N}$ as shown in Fig.~\ref{fig:fig1} (a), making it difficult to generate a correct label or a high confidence score on $x_{gt}$ using Eq.~\ref{eqn:eq_base}. As shown in Fig.~\ref{fig:fig1} (b), with Eq.~\ref{eqn:eq_base} the noisy logits result in many samples correctly classified with low confidence scores or misclassified with high confidence scores (blue curve). Moreover, the long-tail phenomena that inherently occur in a large vocabulary make it hard for a model to well learn effective classifier and feature representations against tail classes. To alleviate the above problems, we propose a classification forest (\cref{subsec:classification_tree} and \cref{subsec:classification_forest}) for enhancing the capability of classifying a large number of categories and an NMS Resampling (\cref{subsec:nms_resampling}) for re-balancing the long-tailed data distribution, respectively.

\begin{figure}
	\centering
	\includegraphics[width=1\linewidth]{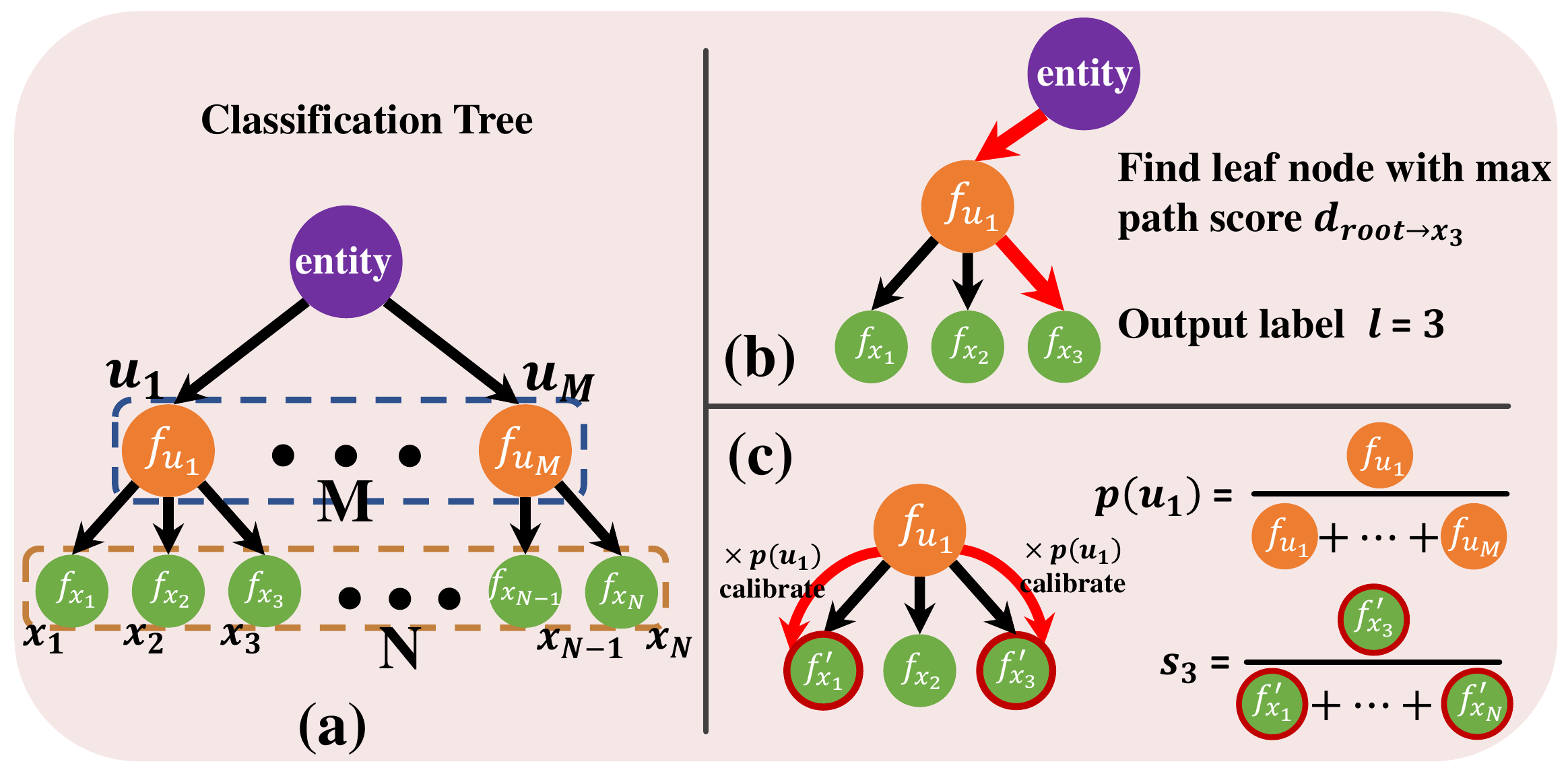}
	\vspace{-6mm}
	\caption{ (a) Classification tree where parent and fine-grained class nodes are contained in \protect\includegraphics[scale=0.13, trim=0 0.5cm 0 0]{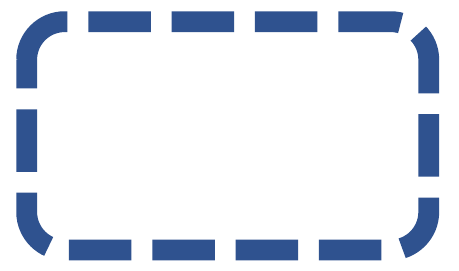} and \protect\includegraphics[scale=0.13,trim=0 0.5cm 0 0]{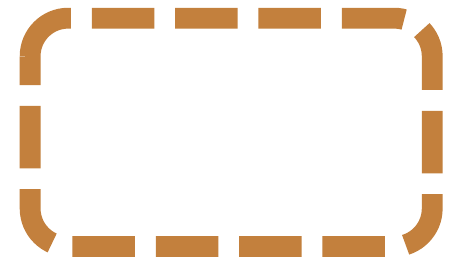}, respectively. \protect\includegraphics[scale=0.25, trim=0 0.4cm 0 0]{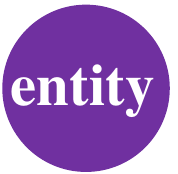} and \protect\includegraphics[scale=0.275, trim=0 0.4cm 0 0]{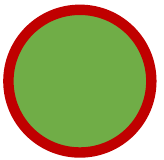} indicates the root node and calibrated leaf node, respectively.	(b) The category label is estimated by finding the maximum path score from the root node to leaf nodes. (c) The confidence scores of fine-grained classes are produced in terms of the values of calibrated leaf nodes.}
	\label{fig:tree}
\vspace{-4mm}
\end{figure}

\subsection{Classification Tree}
\label{subsec:classification_tree}
\myparagraph{Tree Structure.} In real-world scenes, we are aware of prior knowledge of the relationships among object categories. For example, a ``school bus'' and a ``sedan'' have the same parent class ``vehicle'' when considering lexical relation, while ``steering wheel'' and ``basketball'' have the same parent class ``circularity'' when considering geometrical relation. Accordingly, for each type of prior knowledge, all $N$ fine-grained classes can be clustered into $M$ parent classes, where each parent class is on behalf of a characteristic of its children classes. Note that $M$ may vary when we consider different types of prior knowledge. Based on the hierarchical affiliation obtained by prior knowledge, we can build up a tree structure which is a basic component of the classification forest. As shown in Fig.~\ref{fig:tree} (a), the proposed classification tree has 3 levels: the first level is a single root node; the second level consists of $M$ parent class nodes $\{{u_{j}}\}_{j=1}^{M}$; the last level comprises $N$ leaf nodes which represent the fine-grained classes. Each node $v$ (excluding the root node) is associated with its corresponding logit $f_{v}=e^{z_{v}}$ as the node value. We add additional fully connected layers within $\mathcal{N}_{\rm cls}$ to generate the parent class logit $f_{u}$. During training, each level is regarded as an individual classifier and supervised by the softmax function with cross-entropy loss for learning $f_{v}$. Since $f_{v}$ is the classifier logit, it essentially reflects the likelihood that an object belongs to class $v$. Namely, the higher $f_{v}$ is, the more confidently the given object belongs to class $v$ (vice versa).  Also, in the proposed tree structure, we define the path score as the product of the values of nodes through which the path passes (excluding the root node). For an instance, the path score from the root node to $x_{3}$ is $d_{root \rightarrow x_{3}} = f_{x_{3}} \times f_{u_{1}}$.

\begin{figure*}
	\centering
	\includegraphics[width=1\linewidth]{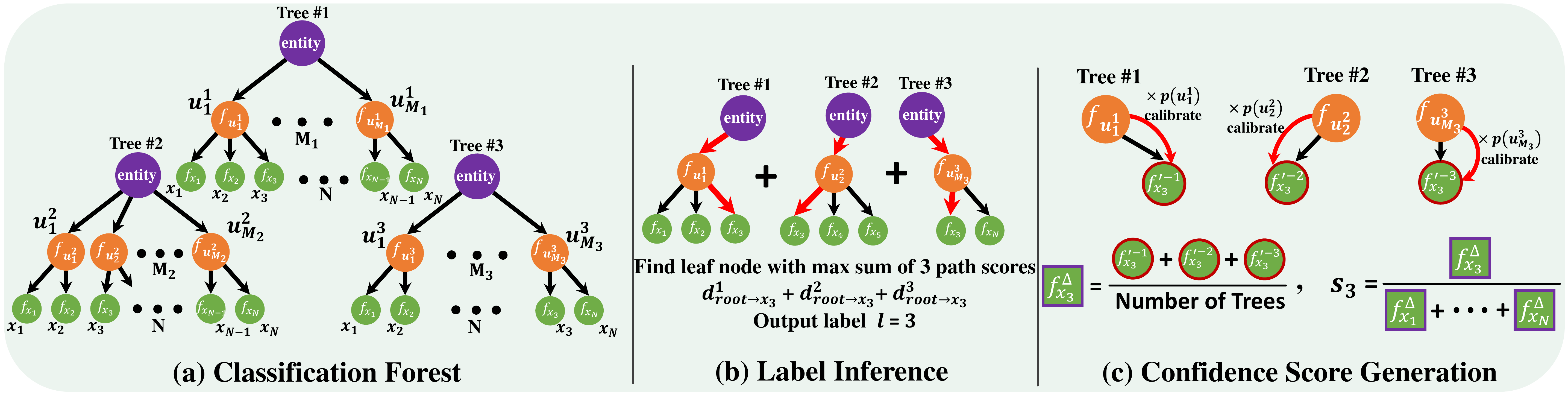}
	\vspace{-6mm}
	\caption{(a) Classification forest with 3 trees. (b) and (c) are brief illustrations of label inference and confidence score generation using (a).}
	\label{fig:forest}
	\vspace{-3mm}
\end{figure*}

\myparagraph{Preliminary Model.} In Eq.~\ref{eqn:eq_base}, both category label and confidence score are determined by the logits of fine-grained classifier only. Different from Eq.~\ref{eqn:eq_base}, our method takes into account the parent classes in order to reduce the negative influence of the noisy logits produced by the fine-grained classifier. To this end, a straightforward way is to combine $p(x_{i})$ with the corresponding $p(u_{j})$ as:

\begin{subequations}
	\begin{align}
	&s_{i}= p(x_{i}) \times p(u_{j}), \quad x_{i}\in ch(u_{j}),\label{eqn:eq2_2} \\
	&l=\underset{i}{\operatorname{argmax}}\ s_{i}, \quad i=1,2,...,N,\label{eqn:eq2_1}
	\end{align}
\label{eqn:eq2}
\end{subequations}
where $ x_{i}\in ch(u_{j})$ indicates the fine-grained class $x_{i}$ that is a child of parent class $u_{j}$. In Eq.~\ref{eqn:eq2}, to parse an object into a fine-grained category, both the fine-grained classifier and parent classifier need to reach a consensus. As an example, the fine-grained classifier wrongly predicts a ``toy'' as a ``sedan'' with $p(x_{sedan})=0.8$, while the probability of the parent class of ``sedan'' is $p(u_{vehicle})=0.05$. In this case, the parent classifier has a different ``opinion'' with the fine-grained classifier, and the final confidence score $s_{toy}$ will be downgraded to $0.04$. Therefore, by incorporating the parent class probability, we are able to reduce the confidence scores of those categories misclassified by the fine-grained classifier, enabling a model to be more fault-tolerant with regard to the noisy logits of the fine-grained classifier.

\myparagraph{Confidence Score Generation.}
In Eq.~\ref{eqn:eq2_2}, the confidence score is generated by the product of fine-grained class probability and parent class probability, which however fails to improve the confidence score of the ground truth class. For example, given a ``sedan'', if $p(x_{sedan})=0.7$ and $p(u_{vehicle})=0.6$, then $s_{sedan}$ becomes $0.42$ unexpectedly. To solve this problem, we use the parent class probabilities to directly calibrate the fine-grained class logits instead of scaling the fine-grained class probabilities. Concretely, the logit/node value of $x_{i}$ is calculated by $f_{x_{i}}^{'}=f_{x_{i}}\times p(u_{j})$, where $ x_{i}\in ch(u_{j})$. Afterwards, similar to the original classifier, the confidence score of the fine-grained class $x_{i}$ is obtained according to the calibrated fine-grained nodes as:

\begin{subequations}
\vspace{-5mm}
	\begin{align}
	s_{i}&=\frac{p(u_{j})\times f_{x_{i}}}{\sum_{a=1}^{N}p(u_{a^{\star}})\times f_{{x_{a}}}}, \quad x_{i}\in ch(u_{j}),\ x_{a}\in ch(u_{a^{\star}}),\label{eqn:eq5_2} \\
	&=\frac{f_{x_{i}}^{'}}{\sum_{a=1}^{N}f_{{x_{a}}}^{'}}.\label{eqn:eq5_1}
	\end{align}
	\label{eqn:eq5}
\vspace{-3mm}
\end{subequations}

Compared to the logits of fine-grained classifier $\{f_{x_{i}}\}_{i=1}^{N}$, $\{f^{'}_{x_{i}}\}_{i=1}^{N}$ are calibrated by the parent class probabilities which can effectively reduce the noisy logits (as evidenced in Fig.~\ref{fig:noisy_logits_comparison}). In contrast to Eq.~\ref{eqn:eq2_2}, Eq.~\ref{eqn:eq5} suppresses the noisy logits on not only negative classes but also the ground truth class. That is to say, we not only reduce the confidence scores of negative classes but also further improve those of the ground truth classes (see Fig.~\ref{fig:fig1} (b)). For the above example of $p(x_{sedan})=0.7$ and $p(u_{vehicle})=0.6$, it indicates that $f_{x_{sedan}}$ account for $70\%$ in $\sum^{N}_{a=1}f_{x_{a}}$. With Eq.~\ref{eqn:eq5}, our method will scale $f_{x_{sedan}}$ by $p(u_{vehicle})=0.6$ and the other $f_{x \notin u_{vehicle}}$ by a smaller $p(u \neq u_{vehicle}) \leq 0.4$. Accordingly, $f_{x_{sedan}}^{'}$ will be less noisy and make up more than $70\%$ of $\sum^{N}_{a=1}f_{x_{a}}^{'}$, which leads to a new $s_{sedan}$ higher than $0.7$. 

\myparagraph{Label Inference.} 
Based on Eq.~\ref{eqn:eq5} and the definition of path score, the category label is inferred by:
\begin{equation}
\label{eqn:eq4}
\begin{aligned}
l&=\underset{i}{\operatorname{argmax}}\ s_{i}\ =\underset{i}{\operatorname{argmax}}\ f^{'}_{x_{i}}, \quad  i=1,2,...,N,\\
&=\underset{i}{\operatorname{argmax}}\ f_{x_{i}} \times f_{u_{j}}, \quad x_{i}\in ch(u_{j}) \ \text{and} \ i=1,2,...,N,\\
&=\underset{i}{\operatorname{argmax}}\ d_{root \rightarrow x_{i}}, \quad i=1,2,...,N.
\end{aligned}
\end{equation}

It is seen that the proposed classification tree parses an object into a fine-grained class by finding the maximum path score from the root node to leaf nodes.

\subsection{Classification Forest}
\label{subsec:classification_forest}
\myparagraph{Forest Structure.} Although a classification tree (\cref{subsec:classification_tree}) can suppress the noisy logits produced by the fine-grained classifier via incorporating the parent class probabilities, false classification can still easily happen if there exist errors in the parent classifier. To alleviate this problem, we build a classification forest that consists of $T$ different classification trees, and each tree will vote for the final classification decision.  As shown in Fig.~\ref{fig:forest} (a), each tree that is based on one type of prior knowledge of category relations may have a different structure from the others, and the parent class nodes set of the $t$-th tree is denoted by $\{u_{j}^{t}\}_{j=1}^{M_{t}}$. The leaf node set, which represents the fine-grained classes of a given dataset, is the same for all the trees. In this paper, we exploit three types of prior knowledge of the relations among fine-grained classes, \emph{i.e.,} lexical relation, visual relation, and geometrical relation, in order to take different aspects of object characteristics into consideration. Based on the relations, we then construct three different classification trees, respectively. Note that the proposed classification forest does not constrain the number of classification trees and we observe that in our experiments three trees are sufficient for achieving improved performance.

\myparagraph{Confidence Score Generation.} In each tree of the classification forest, the value of fine-grained class node $x_{i}$ is calibrated by its corresponding parent class probability as shown in  Fig.~\ref{fig:forest} (c). Thus, for $x_{i}$, we have $T$ different calibrated values $\{f_{x_{i}}^{'-t}\}_{t=1}^{T}$, where $f_{x_{i}}^{'-t}$ implies the likelihood that the $t$-th classification tree considers a given object to belong to $x_{i}$. In order to take advantage of each one of $\{f_{x_{i}}^{'-t}\}_{t=1}^{T}$ and produce a final confidence score $s_{i}$, we take the average of $f_{x_{i}}^{'-1},...,f_{x_{i}}^{'-T}$ and denote it as $f_{x_{i}}^{\bigtriangleup}$, which indicates the comprehensive likelihood that all the classification trees consider a given object to belong to $x_{i}$. Then, similar to Eq.~\ref{eqn:eq5}, the confidence score of $x_{i}$ is calculated according to $\{f_{x_{i}}^{\bigtriangleup}\}_{t=1}^{N}$ as:
\begin{equation}
s_{i}=\frac{f_{x_{i}}^{\bigtriangleup}}{\sum_{a=1}^{N}f_{{x_{a}}}^{\bigtriangleup}}.
\label{eqn:eq7}
\end{equation}

Compared to $f_{x_{i}}^{'-t}$, $f_{x_{i}}^{\bigtriangleup}$ is calibrated by multiple parent classifiers. Since every tree is dedicated to generating confidence score and the likelihood that multiple different parent classifiers yield errors simultaneously is low, the classification forest is robuster than a single classification tree as shown in Tab.\ref{tab:classifcation_forest}. 

\myparagraph{Label Inference.} In \cref{subsec:classification_tree}, the category label is inferred by finding the maximum path score from the root node to leaf nodes. By contrast, to integrate the decisions made by all the trees, we consider a leaf node $x_{i}$ as the predicted category if the sum of each path score from the root node to $x_{i}$ of all the trees is maximum. Formally, the category label is inferred as:
\begin{equation}
\label{eqn:eq6}
l=\underset{i}{\operatorname{argmax}}\ \sum_{t=1}^{T} d_{root \rightarrow x_{i}}^{t}, \quad i=1,2,...,N,
\end{equation}
where $d_{root \rightarrow x_{i}}^{t}$ denotes the path score from the root node to $x_{i}$ in the $t$-th tree. In essence, Eq.~\ref{eqn:eq6} is a plurality vote of all the classification trees, which is generally considered to be better than using a single classification tree as in Eq.~\ref{eqn:eq4}.

\begin{figure}
	\centering
	\includegraphics[width=1\linewidth]{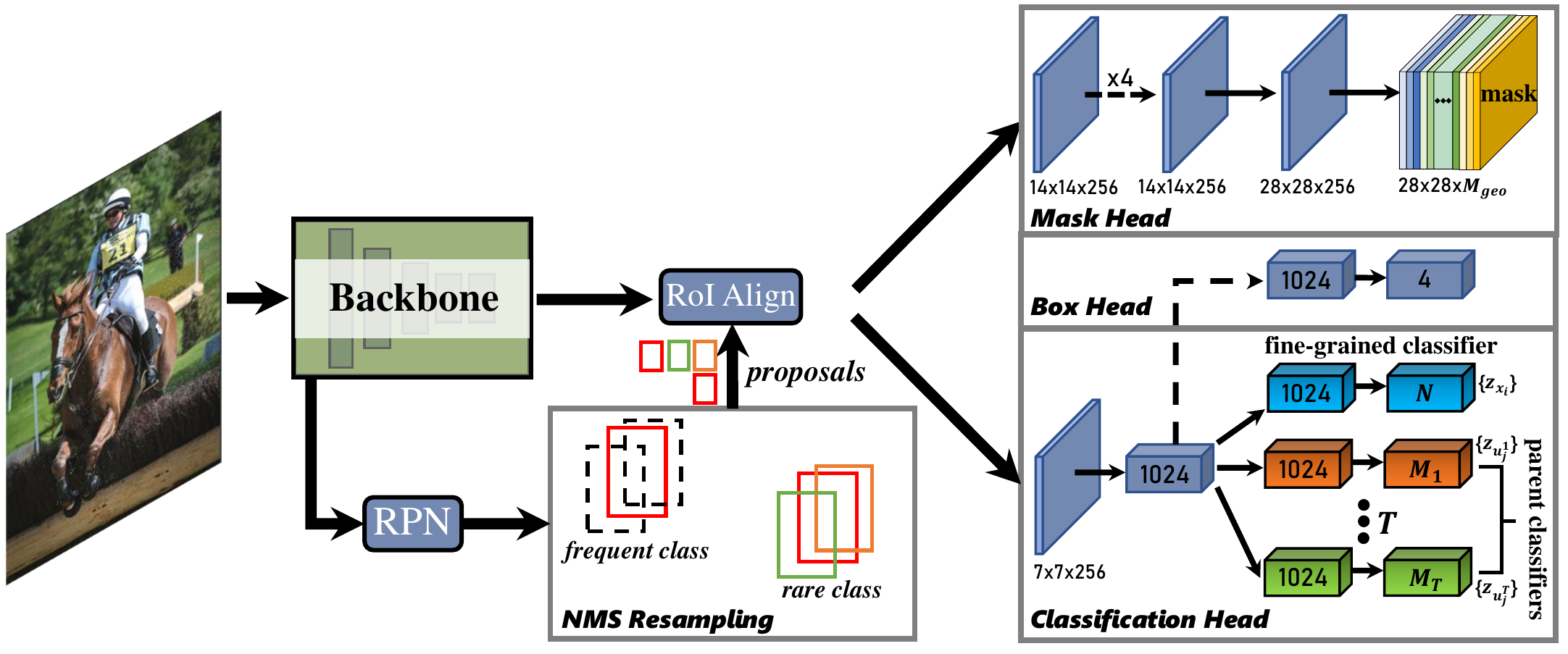}
	\vspace{-4mm}
	\caption{Network architecture of the Forest R-CNN. The dotted rectangles in the NMS Resampling denote the proposal boxes which are suppressed during the NMS process. The NMS Resampling is only used in the training phase.}
	\label{fig:network_architecture}
	\vspace{-3mm}
\end{figure}

\begin{table*}
	\centering
	\caption{Performance comparison with the baseline Mask R-CNN~\cite{maskrcnn} using different backbone networks on the LVIS v0.5 \texttt{val} set. AP denotes the mask AP and AP$^{b}$ denotes the box AP. The subscripts ``r'', ``c'', and ``f'' denote performance on the rare, common, and frequent classes, respectively.}
	\vspace{-4mm}
	\label{tab:tab1}
	\setlength{\tabcolsep}{4pt}
	\begin{tabular}{c|c|c c c|c c c|ccc|ccc}
		Backbone& Method  & AP & AP\textsubscript{50} & AP\textsubscript{75}& AP\textsubscript{\textit{r}} & AP\textsubscript{\textit{c}} & AP\textsubscript{\textit{f}}& AP$^{b}$&AP$^{b}_{50}$&AP$^{b}_{75}$&AP$^{b}_{r}$&AP$^{b}_{c}$&AP$^{b}_{f}$\\
		\toprule[1.3pt]
		
		\multirow{2}*{ResNet-50-FPN} & Mask R-CNN &  21.7 &34.7 &22.8 & 6.8 & 22.6 & 26.4 & 21.8&37.1&22.5&6.5&21.6&28.0  \\
		& Forest R-CNN(Ours)&\bf 25.6 &\bf 40.3 &\bf 27.1 &\bf 18.3 &\bf 26.4 &\bf 27.6 &\bf 25.9&\bf 42.7&\bf 27.2&\bf16.9&\bf26.1&\bf29.2 \\
		
		\hline
		\multirow{2}*{ResNet-101-FPN} & Mask R-CNN & 23.6 &37.1  &25.0  & 10.0 & 24.8 & 27.6 & 23.5&39.9&24.4&8.7&23.1&29.8  \\
		&Forest R-CNN(Ours)&\bf 26.9 &\bf 42.2 &\bf 28.4 &\bf 20.1 &\bf 27.9 &\bf 28.3&\bf27.5&\bf 44.9&\bf 29.0&\bf20.0&\bf27.5&\bf30.4  \\	
		\hline
		\multirow{2}*{ResNeXt-101-32$\times$4d-FPN} & Mask R-CNN &24.8&38.5&26.2&10.0&26.4&28.6&24.8&41.5&25.8&8.6&25.0&30.9\\
		&Forest R-CNN(Ours)&  \bf 28.5 & \bf 43.8 & \bf 30.9 & \bf 21.6 & \bf 29.7 & \bf 29.7 & \bf 28.8&\bf 46.3&\bf 30.9&\bf 20.6&\bf 29.2&\bf 31.7  \\
	\end{tabular}
\vspace{-2mm}
\end{table*}

\subsection{NMS Resampling}
\label{subsec:nms_resampling}
The long-tail phenomena inherently occur in a large vocabulary dataset and the real visual world, in which few classes appear very often but most other classes rarely appear. Such imbalanced data distribution introduces great challenges for learning effective classifier and feature representations against tail classes. To re-balance the long-tailed data distribution, we propose a simple yet effective resampling method, termed as NMS Resampling, by adaptively adjusting the NMS threshold during training. 

As known, after generating a large amount of proposal boxes from the RPN, the NMS is applied to filter out highly overlapped proposals so as to reduce redundancy. The NMS threshold is class-agnostic and set to a fixed value (\emph{e.g.,} $0.7$) for all categories in~\cite{faster_rcnn,maskrcnn}, while our method adaptively adjusts the thresholds for different categories. Specifically, we strive to re-balance the data distribution by utilizing an NMS threshold that is inverse to the data amount of a certain category. That is, we set higher thresholds for tail classes but lower thresholds for head classes. Following this idea, we propose two NMS Resampling schemes to determine the thresholds for specific categories as follows:

\myparagraph{NMS Resampling-Discrete.} The LVIS~\cite{lvis} dataset annotates each category with a category frequency indicating the number of images in which the category appears. Following the descending category frequency, in ~\cite{lvis}, all $1230$ classes are uniformly divided into three groups: \emph{frequent}, \emph{common}, and \emph{rare}. In NMS Resampling-Discrete, we employ three discrete NMS thresholds: $\alpha_{f}$, $\alpha_{c}$, and $\alpha_{r}$ for the \emph{frequent}, \emph{common}, and \emph{rare} classes, respectively, where $\alpha_{f}<\alpha_{c}<\alpha_{r}$. In experiments our method is not sensitive to specific values of $\alpha_{f}$, $\alpha_{c}$, and $\alpha_{r}$  so long as $\alpha_{f}$, $\alpha_{c}$, $\alpha_{r}$ follow ascending order

\myparagraph{NMS Resampling-Linear.} We first uniformly divides three intervals with length $\beta$ for the \emph{frequent}, \emph{common}, and \emph{rare} classes, respectively. Then, in each interval, we linearly assign the NMS threshold to each category as:
\begin{equation}
threshold = 
\begin{cases}
\alpha_{r}+\beta\times\frac{cf_{max}^{f}-cf_{x_{i}}}{cf_{max}^{f}-cf_{min}^{f}}, & \text{if } x_{i} \in frequent \\
\alpha_{c}+\beta\times\frac{cf_{max}^{c}-cf_{x_{i}}}{cf_{max}^{c}-cf_{min}^{c}},              & \text{if } x_{i} \in common\\
\alpha_{f}+\beta\times\frac{cf_{max}^{r}-cf_{x_{i}}}{cf_{max}^{r}-cf_{min}^{r}},              & \text{if } x_{i} \in rare
\end{cases},
\label{eq:grad}
\end{equation}
where $\alpha_{f}<\alpha_{c}<\alpha_{r}$ and $cf_{x_{i}}$ is the category frequency of $x_{i}$. $cf_{max}^{f}$ and $cf_{min}^{f}$ are the maximum and minimum category frequencies in the $frequent$ class group. $\alpha_{f}$, $\alpha_{c}$, $\alpha_{r}$, and $\beta$ are respectively set to $0.65$, $0.75$, $0.85$, and $0.1$ by default in our experiments.

Given a foreground proposal box $\x{b}_{p}$, we first compute its corresponding NMS threshold based on the above schemes. Then, the remaining proposal boxes will be suppressed if they have overlaps with $\x{b}_{p}$ large then the threshold, otherwise, they will be preserved for the next round of NMS procedure. We use the original NMS threshold of $0.7$ for background proposals. The proposed NMS Resampling eases the problem of imbalanced data distribution by preserving more training proposal candidates from the tail classes and suppressing some of those from head classes during training. Compared with the image resampling \cite{lvis}, our method not only is more effective (as shown in Tab.\ref{tab:resample_comparision}) but also avoids repeating training images which may cause overfitting and extra training time. In principle, the NMS Resampling is also applicable to single-stage detectors that need the NMS to reduce redundancy during training, and we leave it for future work.

\subsection{Network Architecture and Loss Function.}
\myparagraph{Network Architecture.} The overall network architecture of the proposed Forest R-CNN is shown in Fig.~\ref{fig:network_architecture}. In the classification head $\mathcal{N}_{\rm cls}$, we add $T$ extra fully connected layer (FC) branches for predicting the node values of parent classes compared with~\cite{maskrcnn}. Since the extra FC branches are inserted after the first FC of $\mathcal{N}_{\rm cls}$ whose channel dimensions are $1024$, our method only introduces little additional computational overhead. Moreover, we incorporate the prior knowledge of geometrical relation into the mask head $\mathcal{N}_{\rm mask}$. In contrast to the original class-specific mask head, we reduce the number of output channels from $N$ to $M_{geo}$, where $M_{geo}$ is the number of parent classes in the geometrical tree. In inference, the mask of class $x_{i}$ is fetched from the $j$-th channel, where $x_{i}\in ch(u_{j}^{geo})$ and $u_{j}^{geo}$ is the $j$-th parent class in the geometrical tree.

\myparagraph{Loss Function.} The overall loss function of the Forest R-CNN is defined as:
\begin{equation}
L=L_{MR}+L_{cls-p}^{1}+...+L_{cls-p}^{T},
\end{equation}
where $L_{MR}$ is the original loss of Mask R-CNN and $L_{cls-p}^{t}$ is the parent classification loss of the $t$-th classification tree.

\section{Experiments}
\subsection{Experiment Setup}

\myparagraph{LVIS Dataset.} LVIS~\cite{lvis} is a large vocabulary dataset for object detection and instance segmentation. There are in total $1230$ and $1203$ categories in LVIS v0.5 and v1.0 datasets, which follow a long-tailed data distribution. All categories are divided into three groups based on the number of images that contains those categories: \emph{rare} (1-10 images), \emph{common} (11-100 images), and \emph{frequent} (>100 images). Our method is trained on the \texttt{train} set and evaluated on the \texttt{val} set. We adopt the evaluation metric AP across IoU threshold from 0.5 to 0.95 for both object detection and instance segmentation results. Our major experiments and ablation studies are performed on LVIS v0.5 dataset. We update the results of Forest R-CNN on LVIS v1.0 dataset in this arxiv V2 version, which is shown in~Tab.\ref{tab:v1}.

\begin{figure*}
	\centering
	\includegraphics[width=1\linewidth]{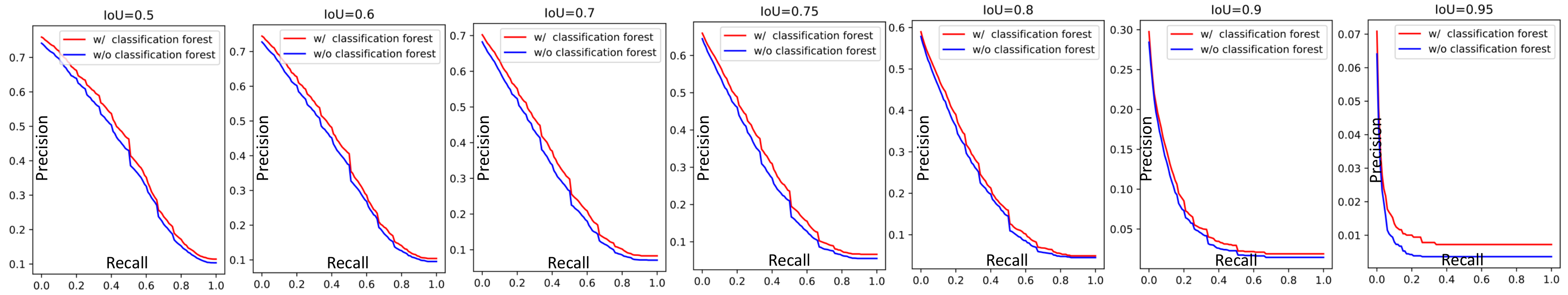}
	\vspace{-7mm}
	\caption{The mask precision-recall (PR) curves w/ and w/o using the classification forest under different IoU thresholds.}
	\vspace{-3mm}
	\label{fig:pr_cureve}
\end{figure*}

\myparagraph{Implementation Details.} The Forest R-CNN uses the same basic settings as in~\cite{lvis}, \emph{e.g.,} image size, score threshold, and the number of object instances per image. Our method is trained with $24$ epochs using the SGD optimizer, and the initial learning rate is set to $0.02$ and decreased by a factor of $10$ at the $16$-th epoch and $22$-th epoch, respectively. We use three relations among fine-grained categories, \emph{i.e.,} lexical relation, visual relation, and geometrical relation, respectively, to construct three classification trees before training. For the visual and geometrical trees, we employ K-means to cluster the visual features and ground truth binary masks of $1230$ fine-grained categories into $M_{vis}$ and $M_{geo}$ parent classes, respectively, where the visual features are obtained by the Mask-RCNN baseline. $M_{vis}$ and $M_{geo}$ are set to $25$ and $50$, respectively. For the lexical tree, it is constructed according to a subgraph of WordNet~\cite{wordnet}, which results in $M_{lex}=108$ parent classes. In the following experiments, our method is equipped with the NMS Resampling-Discrete and ResNet-50-FPN by default. For the NMS Resampling-Discrete, we empirically set $\alpha_{f}$, $\alpha_{c}$, $\alpha_{r}$ to $0.7$, $0.8$ and $0.9$ by default.

\begin{table}
	\centering
	\caption{Ablation study of the proposed NMS Resampling and classification forest. ``NR'' denotes the NMS Resampling.}
	\vspace{-3mm}
	\label{tab:tab2}
	\begin{tabular}{cc|c c| c c c}
		NR& Classification Forest  & AP& AP$^{b}$ & AP\textsubscript{\textit{r}} & AP\textsubscript{\textit{c}} & AP\textsubscript{\textit{f}}\\
		\toprule[1.3pt]
		&&21.7&21.8&6.8&22.6&26.4\\
		\Checkmark&&23.5&23.5&15.6&24.1&25.9\\
		&\Checkmark&23.6&24.0&10.9&24.4&27.5\\
		\Checkmark&\Checkmark&\bf25.6&\bf25.9&\bf18.3&\bf26.4&\bf27.6 	
	\end{tabular}
\vspace{-3mm}
\end{table}

\subsection{Ablation Studies}

\myparagraph{Comparison with Baselines.} As shown in Tab.\ref{tab:tab1}, we compare the proposed Forest R-CNN with the baseline Mask R-CNN under different backbone networks. The Forest R-CNN consistently improves AP and AP$^{b}$ over the baseline with significant gains of $10.1\%$-$12\%$ and $3.3\%$-$4.1\%$ on rare categories and overall categories, respectively. We also separately assess the proposed classification forest (\cref{subsec:classification_forest}) and NMS Resampling (\cref{subsec:nms_resampling}) in Tab.\ref{tab:tab2}. We see from the results that the NMS Resampling improves AP$_{r}$ from $6.8\%$ to $15.6\%$ with the ResNet-50~\cite{resnet}, demonstrating the strong effectiveness of our re-balancing schemes for tail classes. With the classification forest, our method consistently improves performance on overall categories, which shows that the classification forest is specialized to recognize a large number of categories. Visualized samples can be found in Fig.~\ref{fig:example}.

\begin{figure}
	\centering
	\includegraphics[width=1\linewidth]{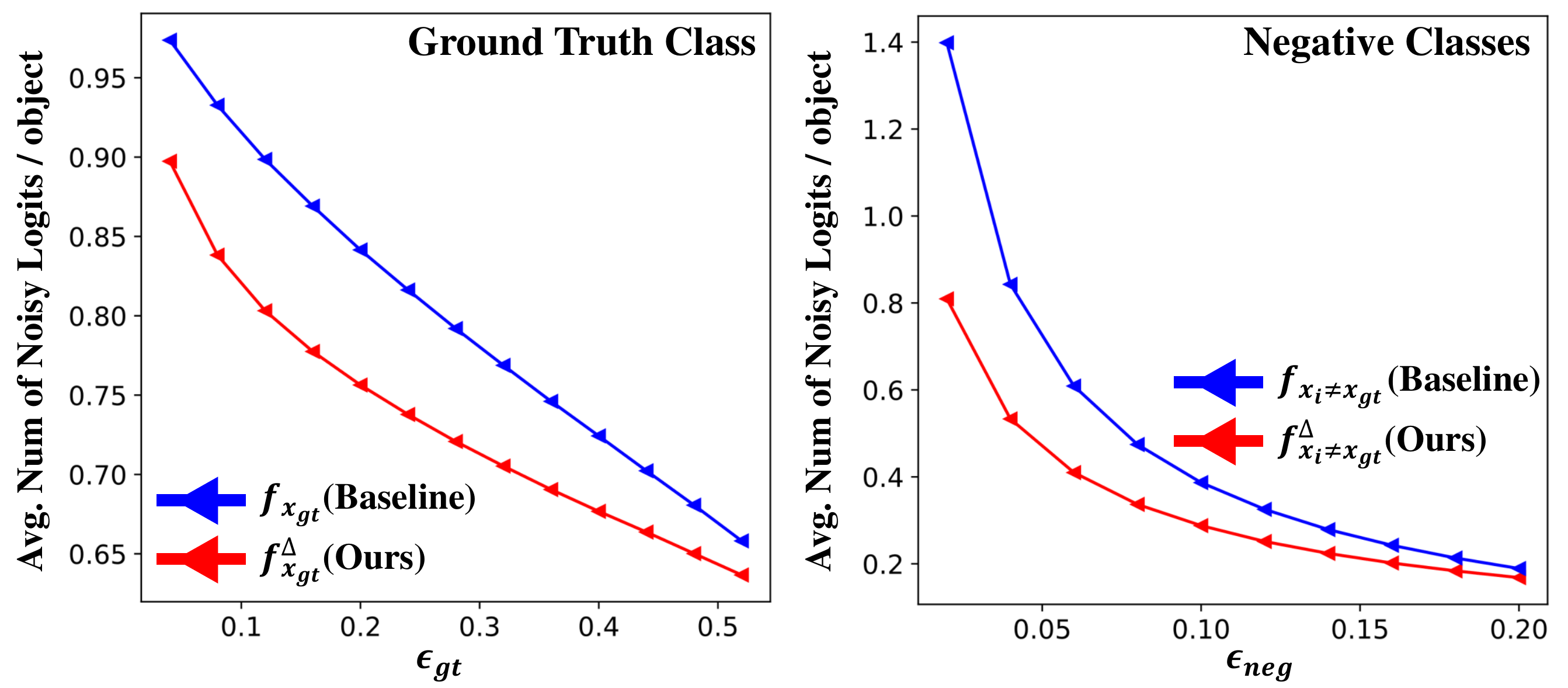}
	\vspace{-5mm}
	\caption{Statistics of the noisy logits w/ and w/o using the classification forest on the LVIS v0.5 \texttt{val} set. The results suggest that the proposed classification forest can effectively suppress the noisy logits from the fine-grained classifier. The NMS resampling is enabled for both baseline and ours.}
	\vspace{-1mm}
	\label{fig:noisy_logits_comparison}
\end{figure}

\begin{figure*}
	\centering
	\includegraphics[width=1\linewidth]{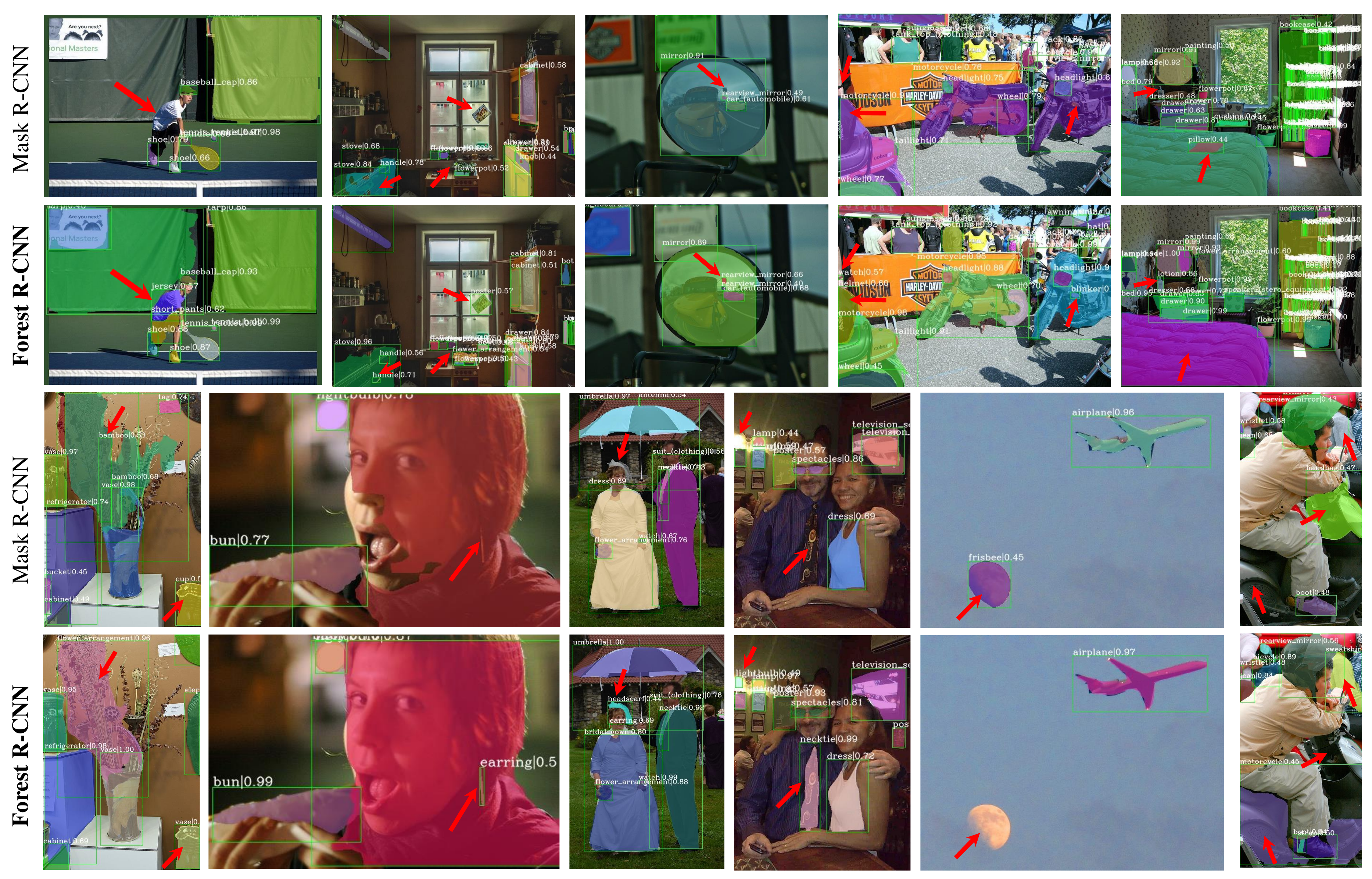}
	\vspace{-6mm}
	\caption{Mask R-CNN~\cite{maskrcnn} \emph{vs.} Forest R-CNN. Mask R-CNN exhibits more wrong classification and miss recognition. For neat visualization, we apply the NMS with threshold of $0.7$ and filter out the predictions with scores lower than $0.4$.}
	\label{fig:example}
\end{figure*}

\myparagraph{Effectiveness of the Classification Forest.} To study the effectiveness of the classification forest (\cref{subsec:classification_forest}), we evaluate the average number of noisy logits per object on the LVIS v0.5 \texttt{val} set. As shown in Fig.~\ref{fig:noisy_logits_comparison}, $\{f_{x_{i}}^{\bigtriangleup}\}_{t=1}^{N}$ contains fewer noisy logitis than $\{f_{x_{i}}\}_{t=1}^{N}$, which demonstrates that the classification forest can effectively suppress the noisy logits produced by the fine-grained classifier. We also investigate the probability density distribution of confidence scores as presented in Fig.~\ref{fig:fig1} (b). We clearly see from the figure that with classification forest our method can effectively improve the confidence scores of correct classified objects and suppress those of wrongly classified objects. This can help to improve the precision/recall of a model under the same recall/precision as evidenced in Fig.~\ref{fig:pr_cureve}. Besides, we evaluate the Forest R-CNN with different settings of the classification trees. As shown in Tab.\ref{tab:classifcation_forest}, the Forest R-CNN with a single classification tree (\cref{subsec:classification_tree}) improves around $1\%$ AP over the baseline, and the lexical tree achieves slightly better performance. With multiple trees, our method further boosts the AP to $24.9\%$-$25.6\%$, validating the effectiveness of the classification forest (\cref{subsec:classification_forest}). 

\begin{table}
	\centering
	\caption{Results of the Forest R-CNN using different classification trees.}
	\vspace{-3mm}
	\label{tab:classifcation_forest}
	\setlength{\tabcolsep}{4pt}
	\begin{tabular}{ccc|cc|ccc}
		Geometrical&Visual&Lexical&AP&AP$^{b}$& AP\textsubscript{\textit{r}} & AP\textsubscript{\textit{c}} & AP\textsubscript{\textit{f}}\\
		\toprule[1.3pt]
		
		& &  &23.5&23.5&15.6 &24.1 &25.9\\
		\Checkmark&&&24.3&24.3&15.2&24.8&27.3\\
		& \Checkmark&&24.3&24.7&14.8 & 25.0 & 27.2\\	
		& &\Checkmark&24.6&24.9&16.5&25.0&27.2\\
		&\Checkmark&\Checkmark&24.9&25.3&16.0&25.8&27.4\\	
		\Checkmark&\Checkmark&&25.2&25.4&17.5&25.8&27.5\\	
		\Checkmark&\Checkmark&\Checkmark&\bf25.6&\bf25.9&\bf18.3&\bf26.4 &\bf27.6
	\end{tabular}
\vspace{-3mm}
\end{table}

\begin{table}
	\centering
	\caption{Performance comparison with the image resampling~\cite{lvis}. ``NR-Linear'' denotes the NMS Resampling-Linear and ``IR'' denotes the image resampling.}
	\vspace{-3mm}
	\label{tab:resample_comparision}
	\begin{tabular}{cc|c c| c c c}
		IR~\cite{lvis}& NR & AP& AP$^{b}$ & AP\textsubscript{\textit{r}} & AP\textsubscript{\textit{c}} & AP\textsubscript{\textit{f}}\\
		\toprule[1.3pt]
		
		&	&  21.7 & 21.8  & 6.8 & 22.6 &\bf 26.4  \\
		\Checkmark & & 23.0&22.7  &13.8  &23.4  &26.1 \\
		
		& \Checkmark({\small NR-Linear}) & \bf 23.7&\bf23.5&12.3 &\bf25.2& 26.3 \\
		
		& \Checkmark({\small NR-Discrete}) &23.5 &\bf23.5&\bf15.6 &24.1 &25.9 \\
		
	\end{tabular}
\vspace{-3mm}
\end{table}

\myparagraph{Effectiveness of the NMS Resampling.} We compare the image resampling method~\cite{lvis} with the two proposed NMS Resampling schemes in~Tab.\ref{tab:resample_comparision}. We see from the results that both the NMS Resampling-Linear and NMS Resampling-Discrete achieve better performance than the image resampling. The reason for the performance gap is that image resampling may introduce severer overfitting by repeating the same images during training. Also, we assess the proposed NMS Resampling with different threshold settings. As shown in~Tab.\ref{tab:resample_ablation}, our method is not sensitive to the specific threshold values so long as they are inverse to the data amount of categories.

\begin{table}
	\centering
	\caption{Results of the NMS Resampling-Discrete with different thresholds.}
	\vspace{-3mm}
	\label{tab:resample_ablation}
	\setlength{\tabcolsep}{4pt}
	\begin{tabular}{c|cc|cccc}
		NMS Resampling-Discrete&AP&AP$^{b}$& AP\textsubscript{\textit{r}} & AP\textsubscript{\textit{c}} & AP\textsubscript{\textit{f}}\\
		\toprule[1.3pt]
		
		$\alpha_{f}=0.7, \alpha_{c}=0.8,\alpha_{r}=0.9$&\bf23.5 &\bf23.5&\bf15.6 &\bf24.1 &25.9 \\
	    $\alpha_{f}=0.6, \alpha_{c}=0.7,\alpha_{r}=0.8$&23.4&23.1&15.2 & 23.8 & \bf26.3
	\end{tabular}
	\vspace{-3mm}
\end{table}

\begin{table}
	\centering
	\caption{Results of the Forest R-CNN using different number of parent classes. The visual tree is used for experiments.}
	\vspace{-3mm}
	\label{tab:parent_classes_number}
	\setlength{\tabcolsep}{4pt}
	\begin{tabular}{c|cc|cccc}
		Number of Parent Classes&AP&AP$^{b}$& AP\textsubscript{\textit{r}} & AP\textsubscript{\textit{c}} & AP\textsubscript{\textit{f}}\\
		\toprule[1.3pt]
		
		$M_{vis}$=25&\bf24.3&24.7&14.8 &\bf25.0 &\bf27.2\\
		$M_{vis}$=50&24.1&\bf24.8&\bf15.0 & 24.8 & 26.8\\
		$M_{vis}$=100&23.7&24.3&12.8 & 24.9 & 26.7\\	
	\end{tabular}
\vspace{-3mm}
\end{table}

\begin{table}
	\centering
	\caption{Performance Comparison on the LVIS v1.0 \texttt{val} set. ``R'' denotes the ResNet with FPN, and ``Cascade'' denotes the Cascade R-CNN~\cite{cai2018cascade}.}
	\vspace{-3mm}
	\label{tab:v1}
	\setlength{\tabcolsep}{2pt}
	\begin{tabular}{c|c|cc|ccc}
		Method&Setting&AP&AP$^{b}$& AP\textsubscript{\textit{r}} & AP\textsubscript{\textit{c}} & AP\textsubscript{\textit{f}}\\
		\toprule[1.3pt]
		Mask R-CNN&R-101&20.8&21.7&1.4&19.4 &30.9\\
		EQL~\cite{equalization_loss}&R-101&22.9&24.2&3.7&23.6 &30.7\\
		Cascade Mask R-CNN&Cascade R-101&22.6&25.2&2.0&22.0 &32.5\\
		De-confound~\cite{tang2020long}&Cascade R-101&23.5&25.8&5.2&22.7 &32.3\\
		\hline
		\hline
		Mask R-CNN&R-50&19.2&20.0&0.0&17.2 &\bf 29.5\\
		EQL~\cite{equalization_loss}&R-50&21.6&22.5&3.8&21.7 &29.2\\
		Forest R-CNN (Ours)&R-50&\bf23.2&\bf24.6&\bf 14.2&\bf22.7 &27.7
	\end{tabular}
\vspace{-3mm}
\end{table}

\begin{table*}
	\centering
	\caption{Performance comparison with the state-of-the-art methods on the LVIS v0.5 \texttt{val} set. We denote ``IR'' as the image resampling~\cite{lvis} and ``MST'' as the multi-scale training. }
	\vspace{-4mm}
	\label{tab:comparison_sota}
	\begin{tabular}{l|c|c c c|c c c|ccc|c}
	Method& Setting  & AP & AP\textsubscript{50} & AP\textsubscript{75}& AP\textsubscript{\textit{r}} & AP\textsubscript{\textit{c}} & AP\textsubscript{\textit{f}}& AP$_{S}$& AP$_{M}$& AP$_{L}$& AP$^{b}$\\
	\toprule[1.3pt]
	
	Class-aware Sampling~\cite{cls_sample} & ResNet-50-FPN &  18.5 & 31.1 & 18.9 & 7.3 & 19.3 & 21.9&13.3&24.3&30.5 & 18.4  \\
	Repeat Factor Sampling~\cite{lvis} &ResNet-50-FPN &  23.2 &-&-& 13.4 & 23.2 & 27.1&-&-&-&- \\
	Class-balanced Loss~\cite{cls_balance_loss} & ResNet-50-FPN & 20.9 & 33.8 & 22.2 & 8.2 & 21.2 & 25.7&15.6&28.1&35.3 & 21.0  \\
	Focal Loss~\cite{focal_loss}& ResNet-50-FPN & 21.0 & 34.2 & 22.1 & 9.3 & 21.0 & 25.8 &15.6&27.8&35.4& 21.9  \\
	LST\cite{lst}& ResNet-50-FPN & 23.0 & 36.7 & 24.8 & - & - & -&-&-&- & 22.6  \\
	EQL~\cite{equalization_loss}&ResNet-50-FPN & 22.8 & 36.0 & 24.4 & 11.3 & 24.7 & 25.1 &16.3&29.7&38.2& 23.3  \\
	Forest R-CNN (Ours)&ResNet-50-FPN&\bf 25.6 &\bf 40.3 &\bf 27.1 &\bf 18.3 &\bf 26.4 &\bf 27.6& \bf18.5& \bf 32.7& \bf 41.1 &\bf 25.9  \\
	
	\hline	
	EQL~\cite{equalization_loss}& ResNet-101-FPN& 24.8 & 38.4 & 26.8 & 14.6 & 26.7 &26.4&-&-&- & 25.2 \\
	Forest R-CNN (Ours)&ResNet-101-FPN& \bf 26.9 &\bf 42.2  &\bf 28.4 &\bf 20.1&\bf 27.9 &\bf 28.3 &\bf 23.6&\bf 43.5&\bf 51.8&\bf 27.5  \\
	
	\hline	
	SOLOv2~\cite{solov2}& ResNet-50-FPN \& MST \& IR& 25.5 & - & - & 13.4 & 26.6 &\bf 28.9&15.9&34.6&\bf 44.9 & - \\
	Forest R-CNN (Ours)&ResNet-50-FPN \& MST& \bf 26.7 &\bf 42  &\bf 28.8 &\bf 19.7&\bf 27.5 & 28.5 &\bf20.3&\bf34.8&39.7&\bf 27  \\
	
	\hline
	SOLOv2~\cite{solov2}&ResNet-101-FPN \& MST \& IR& 26.8 & - & - & 16.3 & 27.6 &\bf 30.1&16.8&35.8&\bf 47.0 & -  \\
	Forest R-CNN (Ours)&ResNet-101-FPN \& MST&\bf 28.2  &\bf 44.3 &\bf 29.7&\bf 20.2  &\bf 29.6  & 29.6&\bf 21.0&\bf 36.0 & 42.0&\bf 28.6  \\
	
	\end{tabular}
\vspace{-4mm}
\end{table*}

\myparagraph{Number of the Parent Classes.} To investigate the impact of hyper-parameter $M$ on the classification tree, we experiment the Forest R-CNN with the Visual tree in Tab.\ref{tab:parent_classes_number}. We see from the table that the performances of $M_{vis}=25$ and $M_{vis}=50$ are close and the results of $M_{vis}=100$ get slightly worse. This suggests that it is better to set $M \leq 50$ when clustering $\sim1000$ fine-grained class into $M$ parent classes using K-means. We observe in experiments that the results are stable under different K-means initializations.

\subsection{Comparison with State-of-the-Art}
\myparagraph{LVIS v0.5.} In Tab.\ref{tab:comparison_sota}, we compare our method with state-of-the-art methods on the LIVS v0.5 dataset. It is worth noting that the proposed Forest R-CNN achieves state-of-the-art performance under different experimental setups. Moreover, the Forest R-CNN improves $3.9\%$-$7.0\%$ AP over the second-best results of different setups on the rare category. It demonstrates that our method is skilled in recognizing the tail classes as well.

\myparagraph{LVIS v1.0.} We report the result of Forest R-CNN with ResNet-50-FPN on the LVIS v1.0 dataset in Tab.\ref{tab:v1}. Forest R-CNN improves $4\%$ AP and $14.2\%$ AP on overall categories and rare categories, respectively, compared to the baseline Mask R-CNN. Moreover, Forest R-CNN achieves competitive performance compared to EQL~\cite{equalization_loss} and De-confound~\cite{tang2020long} which are equipped with more complex network settings.

\section{Conclusion}
This work presents a novel object recognition model, Forest R-CNN, which is equipped with two key components: (\textbf{i}) the classification forest and (\textbf{ii}) the NMS Resampling. The classification forest suppresses the noisy logits produced by a fine-grained classifier, enhancing the capability of classifying thousands of categories. The NMS Resampling re-balances the long-tailed data distribution by adaptively adjusting the NMS thresholds for different categories, which aids our method in recognizing more objects from tail classes. The above designs enable strong performance on detecting and segmenting a large number of object instances, outperforming state-of-the-art competitors on the LVIS dataset.

\myparagraph{Acknowledgment.} This work is supported in part by the start-up funds from State University of New York at Buffalo and gift grant from Horizon Robotics.

\bibliographystyle{ACM-Reference-Format}
\bibliography{Forest_RCNN}

\end{document}